\pdfoutput=1

\documentclass[11pt]{article}

\usepackage[preprint]{acl}

\usepackage{times}
\usepackage{latexsym}
\usepackage{hyperref}
\usepackage{url}
\usepackage{inconsolata}
\usepackage{amssymb}
\usepackage{amsmath}
\usepackage{xspace}
\usepackage{graphicx}
\usepackage{booktabs}
\usepackage{multirow}
\usepackage{caption}
\usepackage{subcaption}
\usepackage[normalem]{ulem} 
\usepackage{todonotes}
\usepackage{enumitem}
\usepackage{colortbl}

\usepackage[T1]{fontenc}

\usepackage[utf8]{inputenc}

\usepackage{microtype}

\usepackage{inconsolata}

\usepackage{graphicx}
\usepackage{extpfeil}
\usepackage{soul}
\usepackage{xcolor}
\usepackage{color}
\usepackage{adjustbox}
\usepackage{xurl}

\newcommand{\ours}{\textit{Lookback Lens}\xspace}

\newcommand{\oursdecoding}{\textit{Lookback Lens Guided Decoding}\xspace}
\newcommand{\red}[1]{{\color{red}{#1}}}

\definecolor{oc-violet-8}{HTML}{6741D9}
\definecolor{oc-indigo-0}{HTML}{EDF2FF}

\title{\emph{Lookback Lens}: Detecting and Mitigating Contextual Hallucinations in Large Language Models Using \emph{Only} Attention Maps}

\author{Yung-Sung Chuang$^\dagger$ \hspace{1mm} Linlu Qiu$^\dagger$ \hspace{1mm} Cheng-Yu Hsieh$^\ddagger$ \quad Ranjay Krishna$^\ddagger$  \\
\bf  \quad Yoon Kim$^\dagger$ \quad James Glass$^\dagger$ \\
  Massachusetts Institute of Technology$^\dagger$ \quad
  University of Washington$^\ddagger$ \\
  \texttt{yungsung@mit.edu} \\
    }

\begin{document}
\maketitle

\begin{abstract}
When asked to summarize articles or answer questions given a passage, large language models (LLMs) can hallucinate details and respond with unsubstantiated answers that are inaccurate with respect to the input context.
This paper describes a simple approach for detecting such \emph{contextual hallucinations}. We hypothesize that contextual hallucinations are related to the extent to which an LLM attends to information in the provided context versus its own generations. Based on this intuition, we propose a simple hallucination detection model whose input features are given by the ratio of attention weights on the context versus newly generated tokens (for each attention head).  We find that a linear classifier based on these \textit{lookback ratio} features is as effective as a richer detector that utilizes the entire hidden states of an LLM or a text-based entailment model. 
The lookback ratio-based detector—\textbf{\ours}—is found to transfer across tasks and even models, allowing a detector that is trained on a 7B model to be applied (without retraining) to a larger 13B model.   
We further apply this detector to mitigate contextual hallucinations, and find that a simple classifier-guided decoding approach is able to reduce the amount of hallucination, for example by 9.6\% in the XSum summarization task.\footnote{Source code: \href{https://github.com/voidism/Lookback-Lens}{\texttt{github.com/voidism/Lookback-Lens}}}
\end{abstract}

\section{Introduction}

Despite the utility and impressive capabilities of large language models (LLMs), their tendency to generate hallucinations, i.e., content that deviates from facts or contextually relevant information~\citep{ji2023survey}, presents a significant challenge in their deployment. 
In this work, we focus on the scenarios where the model is provided with the correct facts within the input context but still fails to generate accurate outputs, a phenomenon we term \emph{contextual hallucination}. Despite the simplicity of this setup, LLMs struggle with contextual hallucinations, frequently producing errors in tasks such as summarization and document-based question answering (e.g.,  Table~\ref{tab:dataset_stats}), which can cause serious issues in applications such as retrieval-augmented generation (RAG)~\citep{lewis2020retrieval}, even when correct documents are retrieved.

\begin{figure*}[t!]
\begin{center}
\includegraphics[width=\textwidth]{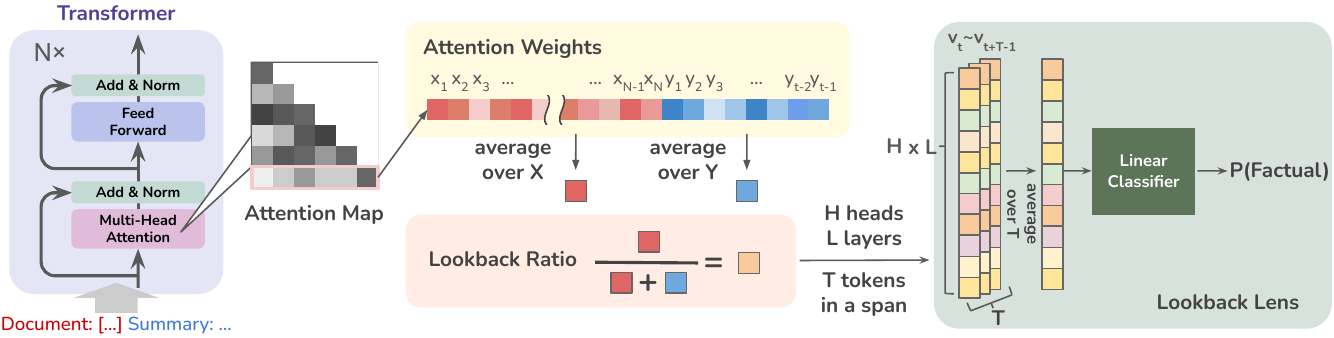}
\end{center}
\caption{An illustration of the \ours. We extract attention weights and calculate the lookback ratios for all layers and all heads. We train a linear classifier on the concatenated features to predict truthfulness of the generation.}
\label{fig:main}
\end{figure*}

Most prior studies that propose methods to combat hallucination focus on the scenario \emph{without} any input context, where the hallucinations arise from the LLMs' parametric knowledge. These works detect and mitigate hallucinations by generally using the LLM's representations, such as hidden states~\citep{burns2022discovering, azaria2023internal}, MLP outputs~\citep{zhang2024truthx, simhi2024constructing}, attention block outputs~\citep{zhang2024truthx, simhi2024constructing} and attention head outputs~\citep{li2024inference, chen2024truth, simhi2024constructing}. In contrast,   the provided contextual information plays a key role in detecting contextual hallucinations. Insofar as attention (more so than other model internals) provides a human-meaningful measure of how much weight is given to the context during generation, this motivates the use of signals from the {attention maps} for hallucination detection and mitigation.

To leverage signals from attention maps, we start by hypothesizing that contextual hallucinations are related to the extent to which an LLM attends to the provided contextual information.
Concretely, we propose a simple feature called \emph{lookback ratio}, which is computed as the ratio of attention weights on the given context versus the newly generated tokens. At each time step, we calculate this lookback ratio for each attention head, and train a linear classifier, which we call the \ours, to detect contextual hallucinations based on the lookback ratio features, as illustrated in Figure~\ref{fig:main}.
The \ours performs on par with, and sometimes even surpasses, more complex feature-based detectors that utilize hidden states from LLMs or text-based entailment models trained on extensively annotated datasets.
We can further integrate this detector during decoding to derive a \oursdecoding strategy which can reduce contextual hallucinations by 9.6\% from LLaMA-2-7B-Chat in the XSum summarization task.
Furthermore, our use of ``higher level'' attention map features makes it possible to transfer the detector across models \emph{without} retraining, allowing a LLaMA2-13B-Chat model to use the same detector that has been trained on LLaMA-2-7B-Chat, and still reduce hallucinations by 3.2\% in XSum. These results collectively highlight the potential of combating contextual hallucination by leveraging the information from attention maps.
\section{Contextual Hallucinations Detection}
\label{sec:detect}

\subsection{Lookback Lens}
To \emph{detect} contextual hallucinations in LLMs, we introduce a lookback ratio, a measure based on the attention distribution of a transformer model. Given a transformer with \( L \) layers, each with \( H \) heads, the model processes an input sequence of context tokens \(X =  \{x_1, x_2, \dots, x_{N}\} \) of length $N$ followed by a set of newly generated tokens \(Y =  \{y_1, y_2, \dots, y_{t-1}\} \) to generate the next token $y_t$.
For time step $t$, and for each head, we calculate the ratio of attention weights focused on the context tokens versus the newly generated tokens.
Formally, for each head \( h \) in layer \( l \), we define:
\[
A^{l,h}_t(\text{context}) = \frac{1}{N} \sum_{i=1}^{N} \alpha^l_{h,i},
\]
\[
A^{l,h}_t(\text{new}) = \frac{1}{t-1} \sum_{j=N+1}^{N+t-1} \alpha^l_{h,j},
\] 
where \( \alpha^l_{h,i} \) and \( \alpha^l_{h,j} \) are softmax-ed attention weights assigned to context tokens \( X \) and new tokens \( Y \) respectively.
The lookback ratio \( \text{LR}^{l,h}_t \) for head \( h \) in layer \( l \) at time step $t$ is then calculated as:
\[
\text{LR}^{l,h}_t = \frac{A^{l,h}_t(\text{context})}{A^{l,h}_t(\text{context}) + A^{l,h}_t(\text{new})}.
\]

To utilize these lookback ratios as input features in detecting hallucinations, we concatenate the lookback ratios across all heads and layers into a feature vector for the time step $t$:
\[
\mathbf{v}_t = [\text{LR}^{1,1}_t, \text{LR}^{1,2}_t, \dots, \text{LR}^{L,H}_t].
\] 
Given a text span of interest $\{y_t, y_{t+1}, ..., y_{t+T-1}\}$, we average the corresponding lookback ratio vectors $\{\mathbf{v}_t, \mathbf{v}_{t+1}, ..., \mathbf{v}_{t+T-1}\}$ into a single vector $\bar{\mathbf{v}}$.
We then employ a logistic regression classifier $\mathcal{F}$ to predict if the span is factual (1) or hallucinated (0) based on the averaged lookback ratio vector.
\[
P( y = 1 | \mathbf{\bar{v}}) = \mathcal{F}(\mathbf{\bar{v}}) = \sigma(\mathbf{w}^\top \mathbf{\bar{v}} + b),
\]
where \(\sigma\) denotes the sigmoid function, \(\mathbf{w}\) is the weight vector, and \(b\) is the bias term of the classifier. 

\begin{table}[t!]
\centering
\small
\begin{tabular}{lcc}
\toprule
\textbf{Dataset} & \textbf{Examples} & \textbf{Correct} \\
\midrule
CNN/DM           & 1000              & 49.6\%                            \\
NQ & 2655              & 67.8\%     \\
\bottomrule
\end{tabular}
\caption{Dataset statistics and GPT-4o evaluation results on responses greedy decoded by LLaMA-2-7B-chat.}
\label{tab:dataset_stats}
\end{table}
\paragraph{Defining Span}

The \ours predicts the probability of hallucinations over spans. We consider two ways to obtain spans for a given sequence: \emph{predefined spans} or \emph{sliding window}.

\textbf{1) Predefined Spans:} 
When the hallucinated and non-hallucinated span annotations are available, we directly train the classifier to differentiate between them. This is a clean setting where all spans are either hallucinated or non-hallucinated.

\textbf{2) Sliding Window:} 
In practice, we do not have any predefined spans during decoding, thus we need a sliding window setup that iterates over all possible spans.
Specifically, we process the sentences into fixed-sized chunks and train the classifier to predict a label of 0 if any hallucinated content exists within a chunk, and 1 otherwise. Here, the annotated data is only used for creating labels, not for the span segmentation. This is more realistic for classifier-guided decoding, but it presents greater challenges because a chunk can contain both hallucinated and non-hallucinated content.

\subsection{Experimental Setup}
\label{sec:detect_setup}

\paragraph{Data} Training the \ours requires labels for hallucinated and non-hallucinated examples. To obtain these examples, we first prompt LLaMA-2-7B-Chat~\citep{touvron2023llama} to greedy decode responses for 1,000 summarization examples from the CNN/DM dataset~\citep{see2017get} and 2,655 QA examples from the Natural Questions~\citep{kwiatkowski2019natural} following the setup of \citet{liu2024lost}. More details are shown in Appendix~\ref{appx:data_create}.
Although being prompted to generate correct responses, the decoded responses will contain both hallucinated and non-hallucinated information as the LLaMA model is still not perfect.
Then, we employed GPT-4o~\citep{openai2024gpt4o} to verify the truthfulness of these responses and provide span-level annotations on hallucinated segments (detailed prompts in Appendix~\ref{appx:gpt4o}). 

Additionally, we performed a pilot study of human annotation on a subset of 70 examples of the summarization task (details in Appendix~\ref{appx:human}), confirming a 97\% consistency rate between GPT-4o annotations and human judgments, and validating the reliability of the automated annotations. 
We show LLaMA-2-7B-Chat's results on both tasks, as evaluated by GPT-4o, in Table~\ref{tab:dataset_stats}.
The results show that the generated summaries from LLaMA-2-7B-Chat still exhibit hallucinations about half of the time, highlighting the challenge of summarization tasks.

\paragraph{Baselines}
We compare our detection method against several baselines:
\textbf{1) Text-based entailment classifier:} We fine-tune the DeBERTa-v3-base~\citep{he2021debertav3} model on the same dataset of CNN/DM and NQ as a natural language entailment (NLI) task. Additionally, we include the results from a state-of-the-art entailment model~\citep{vectara} trained on a huge amount of annotated NLI data (see details in Appendix~\ref{appx:nli}).

\textbf{2) Hidden states-based classifier:} We train classifiers using the same setting as the \ours but used input features from the hidden states of LLaMA-2-7B-Chat from its 24th, 28th, and 32nd layers instead of the lookback ratio. This baseline resembles a broad range of existing methods in the literature~\citep{azaria2023internal, simhi2024constructing}. Our selection of layers followed the findings outlined in \citet{azaria2023internal}, which used layers 32, 28, 24, and 20 of a 32-layer LLM for detecting hallucinations. They find that layers near the 28th layer are most effective (see Table 3 and 4 in \citet{azaria2023internal}). 

We include additional experiments for leveraging multiple layers or all layers in predicting contextual hallucinations in Appendix~\ref{appx:all_layers}, but the results are not significantly better than using the 28th layer.
Some papers suggest attention block outputs could be more useful for detecting hallucinations~\citep{campbell2023localizing, li2024inference}, we include the additional comparative experiments in Appendix~\ref{appx:attn-out}, but the difference between hidden states and attention block outputs is relatively small.

\begin{table*}[h!]
    \centering
    \small
    \begin{tabular}{lcccccccccccccc}
        \toprule
        & & &  \multicolumn{3}{c}{\bf Predefined Span} & \multicolumn{3}{c}{\bf Sliding Window = 8} \\
        \midrule
        \multirow{2}{*}{\bf Method} & \multirow{2}{*}{\bf Source} & \multirow{2}{*}{\bf Target} & \multicolumn{3}{c}{\bf Source $\xrightarrow{\hspace{0.6cm}}$ Target} & \multicolumn{3}{c}{\bf Source $\xrightarrow{\hspace{0.6cm}}$ Target} \\
        \cmidrule(lr){4-5}\cmidrule(lr){6-6}
        \cmidrule(lr){7-8}\cmidrule(lr){9-9}
        \multirow{2}{*}{} & \multirow{2}{*}{} & \multirow{2}{*}{} & \bf Train & \bf Test & \bf Transfer & \bf Train & \bf Test & \bf Transfer \\
        \midrule
        \multicolumn{9}{c}{\it Text based NLI} \\
        \midrule
        SoTA NLI & -- & Sum. & -- & -- & 76.6 & -- & -- & 57.1 \\
        SoTA NLI & -- & QA & -- & -- & 58.6 & -- & -- & 61.8 \\
        NLI (our impl.) & QA & Sum. & -- & -- & 55.1 & -- & -- & 53.0 \\
        NLI (our impl.) & Sum. & QA & -- & -- & 71.0 & -- & -- & 64.9 \\
        \midrule
        \multicolumn{9}{c}{\it Hidden states based} \\
        \midrule
        32nd Layer & QA & Sum. & 100.0 & 89.6 & 79.4 & 99.0 & 97.1 & 56.1 \\
        32nd Layer & Sum. & QA & 100.0 & 82.5 & 81.8 & 97.0 & 94.8 & 59.4 \\
        28th Layer & QA & Sum. & 100.0 & 91.4 & 83.6 & 99.2 & 97.3 & 57.7 \\
        28th Layer & Sum. & QA & 100.0 & 83.3 & 84.7 & 97.2 & 95.2 & 58.8 \\
        24th Layer & QA & Sum. & 100.0 & 92.0 & 81.3 & 99.2 & 97.4 & 58.3 \\
        24th Layer & Sum. & QA & 100.0 & 83.1 & 83.0 & 99.2 & 97.4 & 58.3 \\
        \midrule
        \multicolumn{9}{c}{\it Attention maps based (Ours)} \\
        \midrule
        \ours & QA & Sum. & 98.3 & 91.2 & 85.3 & 88.3 & 87.1 & 66.1 \\
        \ours & Sum. & QA & 97.7 & 88.8 & 82.0 & 86.2 & 85.3 & 66.0 \\
        \bottomrule
    \end{tabular}
    \caption{AUROC of the classification tasks using predefined span segmentation and sliding window (size = 8) on NQ (QA) and CNN/DM (Sum.). The source task scores (Train/Test) are averaged over two-fold validation.}
    \label{tab:classification_results}
    \vspace{-5pt}
\end{table*}

\subsection{Results}

Our results are presented in Table~\ref{tab:classification_results}. We consider both predefined span segmentation and sliding window with a window size of 8. 
We include the two-fold validation setting on the source task and the out-of-domain transfer setting on the target task, with the tasks either question answering (QA) or summarization (Sum.). We find that the \ours achieves slightly better performance than the hidden states-based classifier and significantly outperforms the NLI models (SoTA and our impl.). The advantage of the \ours over the hidden states-based classifier is more significant in the sliding window settings, as shown in the right-hand side of Table~\ref{tab:classification_results}.

Additionally, we observe that the hidden states-based classifier tends to overfit the training sets during the two-fold validation, and present a substantial performance drop when transferred to out-of-domain tasks. In contrast, \ours, while not always fitting the training set perfectly, consistently exhibits better performance when applied to out-of-domain tasks. This contrast highlights the effectiveness and generalizability of the lookback ratio features we extract from the attention maps.

\begin{figure*}[ht!]
\begin{center}
\includegraphics[width=\textwidth]{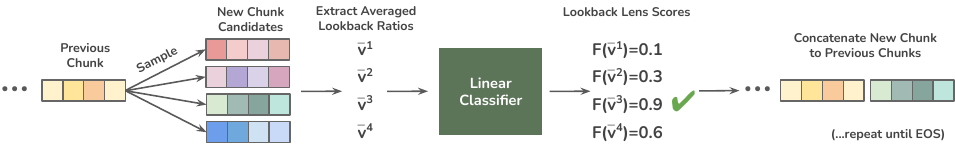}
\end{center}
\caption{\oursdecoding: sample multiple chunk candidates, compute lookback ratios from attention maps to be scored by \ours, and select the best candidate that is less likely to be hallucinations.}
\label{fig:decoding}
\end{figure*}

\section{Contextual Hallucinations Mitigation}
\label{sec:mitigate}

\subsection{Lookback Lens Guided Decoding}

To mitigate the impact of contextual hallucinations identified by the \ours, we introduce a classifier-guided decoding strategy to guide the generation toward more contextually accurate outputs. 
This approach serves as a robustness test of the \ours' ability to handle various text generation scenarios. 
While prior studies on controllable text generation adjust the output probabilities using classifiers based on the output tokens~\citep{yang2021fudge}, our method fundamentally differs by not using the tokens themselves but rather their attention maps during generation.

We propose \oursdecoding, which incorporates \ours(\( \mathcal{F} \)) into the decoding process.
Since all tokens in the vocabulary share the same attention pattern during one decoding step, \( \mathcal{F} \) cannot directly influence one-step token choice. 
Instead, \( \mathcal{F} \) can evaluate multiple-token chunks, as each chunk causes different attention patterns in multiple decoding steps. 

Given the context and partially generated text, we independently sample a set of $k$ candidate chunks \( \{C_1, C_2, \dots, C_k\} \) at the same decoding step $t$.
For each chunk \( C_j \), the associated lookback ratios are averaged to form a feature vector \( \mathbf{\bar{v}}^j \). As shown in Figure~\ref{fig:decoding}, we select the best candidate $C^*$ predicted by \( \mathcal{F} \) and append to the generation,
\[
C^* = \underset{C_j \in \{C_1, C_2, \dots, C_k\}}{\arg\max} \mathcal{F}(\mathbf{\bar{v}}^j).
\]

We repeat this process until it generates the EOS token or reaches the maximum length.

\subsection{Experimental Setup}
We evaluate \oursdecoding on three tasks that involve generating texts conditioned on given contexts, including summarization with XSum~\citep{narayan2018don}, QA with NQ~\citep{kwiatkowski2019natural}, and multi-turn conversations with MT-bench~\citep{zheng2024judging}.

For testing the generalization ability of the \ours, we only train it with the CNN/DM summarization dataset from the detection task in Section~\ref{sec:detect_setup}. Thus, only the XSum dataset will be the same-task transfer setting, while NQ and MT-bench will be cross-task transfer setting.

\paragraph{XSum}
To test the \ours's effectiveness at transferring across data distributions for the same task (summarization), we use 1,000 examples sampled from the testing set of XSum. Prior studies~\citep{maynez2020faithfulness} indicate that traditional evaluation metrics such as ROUGE~\citep{lin2004rouge} or BERTScore~\citep{zhang2019bertscore} correlated poorly with human evaluation on faithfulness and factuality. Recent studies~\citep{chiang2023can, liu2023g} also show a strong correlation between GPT-4~\citep{openai2023gpt4} evaluation and human evaluation. Thus, we report the averaged accuracy from the binary judgments of GPT-4o, with the prompts in Appendix~\ref{appx:gpt4o}. We also conduct a pilot study for human evaluation on GPT-4o's judgment in Appendix~\ref{appx:human}, finding that 97\% of the GPT-4o judgments are consistent with human judgment.

\paragraph{Natural Questions}
We use the NQ data from the setup of \citet{liu2024lost} we describe in Appendix~\ref{appx:data} and evaluate the best span exact match following \citet{kandpal2023large, mallen2023not}. 

\paragraph{MT-Bench} We consider a multi-turn conversations setup where the model needs to follow previous chat history. We use MT-bench~\citep{zheng2024judging}, a multi-turn instruction-following benchmark covering eight categories. We focus exclusively on generating responses for the second turn and use GPT-3.5's responses as the default for the first turn. We use GPT-4 to score the model's answers on a scale of 1 to 10 based on various factors, including \emph{helpfulness, relevance, accuracy, depth, creativity, and level of detail of the response}.

Additionally, since we are particularly interested in mitigating contextual hallucinations, we further exclude math questions and evaluate the remaining 50 general questions. We specifically instruct GPT-4o to focus on whether the responses are faithful to the chat history (see prompt in Appendix~\ref{appx:gpt4o}). We refer to this setup as MT-Bench (hallu.).

\paragraph{Baselines}
To evaluate the performance of our proposed method, we compared it against the following baselines: \textbf{1) Greedy Decoding:} generating responses using the LLaMA-2-7B-Chat model~\citep{touvron2023llama} through greedy decoding. \textbf{2) Other Classifier-Guided Decoding:} using exactly the same setting but with different classifiers introduced in Section~\ref{sec:detect_setup}, including
text-based entailment classifiers and hidden states-based classifiers.

\subsection{Main Results}
\label{sec:main_results}

\begin{table}[t!]
    \centering
    \small
    \begin{tabular}{lcccc}
        \toprule
        \multirow{2}{*}{\bf Method} & \multirow{2}{*}{\bf XSum} & \multirow{2}{*}{\bf NQ} & \multicolumn{2}{c}{\bf MT-Bench} \\
        \cmidrule(lr){4-5}
        &  &   & Hallu. & Ori. \\
        \midrule
        Greedy Decoding & 49.0 & 71.2 & 6.08 & 5.10 \\
        \midrule
        \multicolumn{5}{l}{\it Text-based classifier guided decoding} \\
        \midrule
        SoTA NLI$^\dagger$ & 59.0 & 74.2 & 6.12 & 5.03 \\
        NLI (our impl.) & 44.1 & 72.5 & 5.72 & 4.99\\
        \midrule
        \multicolumn{5}{l}{\it Hidden states based classifier guided decoding} \\
        \midrule
        32nd layer & 48.3 & 73.9 & 5.49 & 4.91\\
        28th layer & 48.9 & 73.0 & 5.71 & 5.06\\
        24th layer & 47.5 & 73.9 & 5.65 & 5.16\\
        \midrule
        \multicolumn{5}{l}{\it \ours guided decoding} \\
        \midrule
        Ours & 58.6 & 74.2 & 6.27 & 5.10 \\
        \bottomrule
    \end{tabular}
    \caption{Decoding results using 8 candidates per chunk in a chunk size of 8. We compare our methods with greedy decoding and classifier-guided decoding using the NLI models, and hidden state representations of different layers. $^\dagger$The SoTA NLI is trained on 731k examples so it may not be directly comparable.}
    \label{tab:performance_comparison}
    \vspace{-10pt}
\end{table}

We show our results using eight candidates per chunk in a chunk size of eight in Table~\ref{tab:performance_comparison}, and the ablation with different chunk sizes is shown in Table~\ref{tab:chunk_size}. \oursdecoding can improve the performance on both in-domain task (XSum, by 9.6\%) and out-of-domain tasks (NQ, by 3\%). The original greedy decoding results on XSum achieved 49.0\% correct
which means 510 examples were hallucinated. Our decoding method  significantly reduced the number of hallucinated examples from 510 to 414, resulting in an 18.8\% reduction in the hallucinated examples. This result is on par with using SoTA NLI to guide the decoding, where SoTA NLI is trained on roughly 731k annotated summarization examples, which is 700$\times$ larger compared to our 1k training set. (See Appendix~\ref{appx:nli}.) In contrast, decoding guided by hidden states-based or the NLI (our implementation) classifiers, both trained on the same data of our method, can only slightly improve the performance on NQ, but not for XSum, probably due to the issue of distribution shift, highlighting the advantages of \ours in generalization ability.

For MT-bench, we evaluate both settings: the original setting (ori.) and the setting that is specifically for judging contextual hallucinations (hallu.). 
We do not expect our method can improve on the original setting, because it evaluates many factors such as helpfulness, relevance, etc. But we expect to see an improvement on the hallucination setting.
The results shown in Table~\ref{tab:performance_comparison} suggest that our decoding method can boost the performance on the hallucination setting while maintaining the same performance in the original setting, which shows that our decoding method is effective in reducing hallucinations without compromising the overall generation quality.

\section{Cross-model Transfer}
\label{sec:cross}

One benefit of using the lookback ratio to capture higher-level model patterns for hallucination detection is its potential to better transfer across models. A classifier trained with one model's lookback ratio could potentially be applied to another model \emph{without} retraining, provided correlation between the target model's attention pattern and that of the original model.
Here, we show that we can transfer a \ours trained on attention maps from LLaMA-2-7B-Chat to LLaMA-2-13B-Chat without any retraining.

\begin{table}[t!]
    \centering
    \small
    \begin{tabular}{cccc}
        \toprule
        \multirow{2}{*}{\bf Source} & \multirow{2}{*}{\bf Target} & \bf Predefined & \bf Sliding \\
        \multirow{2}{*}{} & \multirow{2}{*}{} & \bf Span & \bf Window \\
        \midrule
        \multicolumn{4}{l}{\textit{\ours: Train 13B $\rightarrow$ Test 13B}} \\
        \midrule
        QA & Sum. & 84.0 & 60.4  \\
        Sum. & QA & 84.3 & 60.8  \\
        QA-train & QA & 93.3 & 63.7  \\
        
        \midrule
        \multicolumn{4}{l}{\textit{\ours: Train 7B $\rightarrow$ Test 13B}} \\
        \midrule
        QA & Sum. & 73.5 & 58.8  \\
        Sum. & QA & 78.2 & 60.5  \\
        QA-train & QA & 80.6 & 62.4  \\
        \bottomrule
    \end{tabular}
    \caption{Cross model transfer results on detection tasks.}
    \vspace{-10pt}
    \label{tab:cross-model-detection}
\end{table}

Since the total numbers of attention heads are different in 7B and 13B models, and there is no obvious one-to-one mapping between the heads, we use a linear regression model to map the heads from the 13B model to the heads in 7B model. Concretely, we have 1024 heads in 7B and 1600 heads in 13B. We extract the averaged lookback ratio per head for all the $|D|$ training examples, resulting in a $1024 \times |D|$ matrix and a $1600 \times |D|$ matrix.\footnote{To ensure that two models are generating the same content when extracting lookback ratio, we decode from 7B and run the 13B model on the 7B outputs.}
We then fit a linear regression model to map the heads to reconstruct the 7B heads from 13B heads. After applying the linear transformation to the lookback ratio from 13B, the transformed heads can be directly used by 7B's classifiers. See details in Appendix~\ref{appx:nli}.

The detection results are shown in Table~\ref{tab:cross-model-detection}. We first show the same-model (13B$\rightarrow$13B) + cross-task transfer result, and the cross-model (7B$\rightarrow$13B) + cross-task transfer result. Although cross-model transfer yields slightly worse results compared to same-model transfer, the AUROC scores are still non-trivially high. Consider that doing cross-model + cross-task transfer at the same time may be tough to \ours, we also include one more setting that does training on 2.5K examples of the NQ training set\footnote{The NQ-train 2.5K data is annotated in the same method to annotate NQ testing set, as described in Section~\ref{sec:detect_setup}.} and then transfer to the NQ testing set. We see the cross-model same-task transfer results are even closer to the same-model transfer results.

Given promising results on detection tasks, we apply cross-model transfer to \oursdecoding. We conduct the same-task transfer setting: NQ-train (7B) to NQ (13B), and CNN/DM (7B) to XSum (13B).
In Table~\ref{tab:cross-model-decoding}, we observe a performance improvement similar to same-model transfer using 13B itself, or using the SoTA NLI model applied on the 13B decoding.
However, on cross-task + cross-model transfer settings: CNN/DM (7B) to NQ (13B), we do not observe significant improvements where we attribute to the larger distribution shift. We leave this challenging setting for future work.

\begin{table}[t!]
\centering
\small
\begin{tabular}{lccc}
\toprule
\bf Method & {\bf XSum} & {\bf NQ} \\
\midrule
Greedy & 52.9 & 74.0 \\
\midrule
\multicolumn{3}{l}{\it Text-based classifier guided decoding} \\
\midrule
SoTA NLI$^\dagger$ & 59.6 & 74.4 \\
\midrule
\multirow{2}{*}{\bf Method} & {\bf CNN/DM} & {\bf NQ-train} & {\bf CNN/DM} \\
\multirow{2}{*}{} & {\bf $\rightarrow$XSum} & {\bf $\rightarrow$NQ} & {\bf $\rightarrow$NQ} \\
\midrule
\multicolumn{3}{l}{\it \ours guided decoding} \\
\midrule
13B $\rightarrow$ 13B & 57.9 & 75.6 & 74.8  \\
7B $\rightarrow$ 13B & 56.1 & 76.4 & 73.7 \\
\bottomrule
\end{tabular}
\caption{Cross model transfer from LLaMA-2-7B-chat to LLaMA-2-13B-chat using greedy decoding and classifier guided sampling methods with chunk size 8.}
\label{tab:cross-model-decoding}
\vspace{-5pt}
\end{table}

\section{Discussions and Ablations}
\label{sec:discuss}
In this section, we further conduct various experiments and ablation studies on the \ours and its corresponding classifier guided decoding.

\paragraph{Effect of Chunk Size}
In Section~\ref{sec:main_results} (Table~\ref{tab:performance_comparison}), we experiment with chunk size = 8. Here, we study the effect of varying chunk sizes, from 4, 8, to 16. We see that there is a slight trend that \ours guided decoding prefers shorter chunk size for NQ and longer chunk size for XSum. However, in general the improvements are consistent across different chunk sizes, thus reducing the need to optimize for chunk sizes.

\begin{table}[h!]
    \centering
    \small
    \begin{tabular}{lcccccc}
        \toprule
        \bf Method & \multicolumn{3}{c}{\bf NQ} & \multicolumn{3}{c}{\bf XSum} \\
        \cmidrule(lr){2-4} \cmidrule(lr){5-7}
        Chunk size=  & 4 & 8 & 16 & 4 & 8 & 16 \\
        \midrule
        Greedy & \multicolumn{3}{c}{71.2 } & \multicolumn{3}{c}{49.0} \\
        \midrule
        \multicolumn{7}{l}{\it Text-based classifier guided decoding} \\
        \midrule
        SoTA NLI$^\dagger$ & 73.7 & 74.2 & 74.4 & 57.3 & 59.0 & 62.1 \\
        \midrule
        \multicolumn{7}{l}{\it Hidden states based classifier guided decoding} \\
        \midrule
        32nd layer & 72.6 & 73.9 & 72.7 & 48.9 & 48.3 & 48.3  \\
        28th layer & 72.9 & 73.0 & 74.1 & 47.2  & 48.9 & 47.1 \\
        24th layer & 75.0 & 73.9 & 72.5 & 47.6 & 47.5 & 51.2 \\
        \midrule
        \multicolumn{7}{l}{\it \ours guided decoding} \\
        \midrule
        Ours & 75.4 & 74.2 & 74.3 & 53.2 & 58.6 & 57.7 \\
        \bottomrule
    \end{tabular}
    \caption{Performance comparison on various datasets using different methods and chunk sizes.}
    \label{tab:chunk_size}
    \vspace{-10pt}
\end{table}

\paragraph{Predictive Power of Different Heads}
\begin{table}[ht!]
    \centering
    \small
    \begin{adjustbox}{width=\linewidth}
    \begin{tabular}{lccccccc}
        \toprule
        \multirow{4}{*}{\bf Method} & \multicolumn{6}{c}{\bf Predefined Span} \\
        \cmidrule(lr){2-7}
        & \multicolumn{3}{c}{\bf QA $\rightarrow$ Sum.} & \multicolumn{3}{c}{\bf Sum. $\rightarrow$ QA}  \\
        \midrule
        All heads & \multicolumn{3}{c}{85.3} & \multicolumn{3}{c}{82.0} \\
        \midrule
        \multicolumn{7}{l}{\it Top-$k$ heads only} \\
        \cmidrule(lr){2-4} \cmidrule(lr){5-7}
        \multicolumn{1}{r}{\it with $k =$} & 10 & 50 & 100 & 10 & 50 & 100 \\
        \midrule
        Largest mag. & 71.2 & 82.3 & 82.8 & 79.2 & 80.3 & 81.1 \\
        Most positive & 65.1 & 74.9 & 75.4 & 66.3 & 70.3  & 74.4 \\
        Most negative & 59.5 & 67.5 & 74.4 & 66.4 & 70.2 & 73.0 \\
        \bottomrule
    \end{tabular}
    \end{adjustbox}
    \caption{Cross-task transfer AUROC using top-$k$ attention heads selected according to: coefficients with the largest magnitude (largest mag.), most positive, and most negative. We consider $k$ = 10, 50, and 100.}
    \label{tab:topk}
\end{table}

In the aforementioned experiments, we utilize all attention heads to train the \ours. We are thus interested in 
how the predictive power is distributed among different heads in making predictions. That is, how much performance can we recover if we only utilize a subset of heads?
To answer this, we use the coefficients in the linear classifier of the \ours (in Section \ref{sec:detect}) to estimate the importance of each head in detecting hallucinations.

In Table~\ref{tab:topk}, we show the results on detection tasks achieved by different detectors trained using only a subset of top-$k$ heads with the largest magnitude of coefficients in the original \ours trained will all heads.
The results show that the predictive power is not concentrated only on a subset of heads. 
Using only top-10 heads is worse than using all heads, and increasing $k$ consistently improves performance and top-100 heads largely recover the model's performance using all heads.

More interestingly, we also include the results that only select the top-$k$ heads among the heads with most positive/negative coefficients, which are positive/negatively correlated to factuality.
On the heads with positive coefficients, higher lookback ratio (i.e., when the heads attend at the context more) indicates higher factuality and less hallucination; conversely, heads with negative coefficients suggest a lower lookback ratio (i.e., attending to generated tokens more) is more likely to be truthful.
Table~\ref{tab:topk} shows that none of positive or negative heads alone can be on par with using the top-$k$ largest magnitude heads.
This result implies that both positive and negative heads are critical for a model to generate factual responses. We conjecture that the positive heads may specialize at context grounding, and thus higher lookback ratio on these heads leads to more factual response. On the other hand, the negative heads may be critical at ensuring consistency in its own generation, and thus should attend to the generated tokens more.
We leave further investigation on this interesting balance for future work. Meanwhile, we visualize the lookback ratio of positive/negative heads in Appendix~\ref{appx:vis}. 

\begin{figure*}[th!]
\begin{center}
\includegraphics[width=\textwidth]{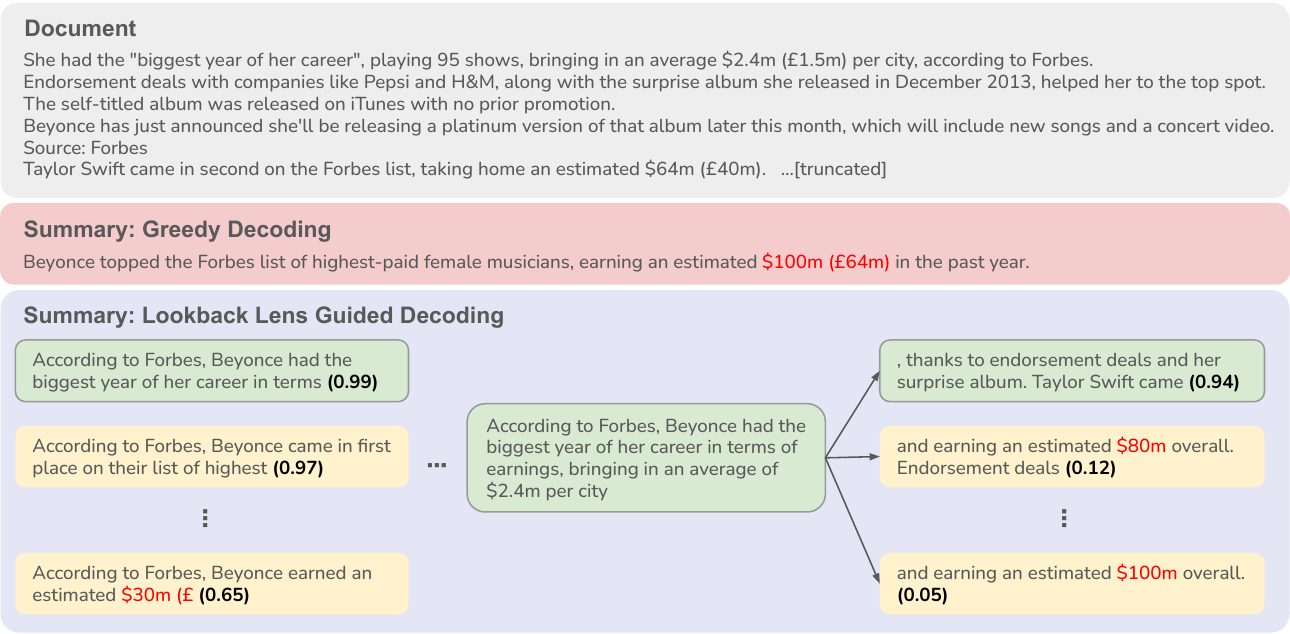}
\end{center}
\vspace{-10pt}
\caption{Qualitative example on XSum using the LLaMA-2-7B-Chat model with greedy decoding and \oursdecoding. The numbers in the parenthesis show the predicted scores from the \ours.}
\label{fig:qual}
\vspace{-10pt}
\end{figure*}

\begin{table}[t!]
    \centering
    \small
    \begin{tabular}{lcc}
        \toprule
        \multirow{3}{*}{\bf Layers} & \multicolumn{2}{c}{\bf Predefined Span} \\
        \cmidrule(lr){2-3}
        & \bf QA $\rightarrow$ Sum. & \bf Sum. $\rightarrow$ QA \\
        \midrule
        Layer 1-4 & 69.6 & 64.0 \\
        Layer 5-8 & 75.6 & 70.1 \\
        Layer 9-12 & 75.4 & 68.3 \\
        Layer 13-16 & 81.2 & 78.2 \\
        Layer 17-20 & 80.8 & 78.2 \\
        Layer 21-24 & 64.4 & 73.1 \\
        Layer 25-28 & 66.0 & 74.4 \\
        Layer 29-32 & 66.4 & 71.4 \\
        \midrule
        Layer 1-32 & 85.3 & 82.0 \\
        \bottomrule
    \end{tabular}
    \caption{Cross-task transfer AUROC among layers.}
    \label{tab:layerwise_transfer_auroc}
    \vspace{-10pt}
\end{table}

\paragraph{Reducing Number of Layers}

We experiment with using only a subset of layers for \ours, as shown in Table~\ref{tab:layerwise_transfer_auroc}. We can see that the predictive power is not concentrated in any subset of layers, as none of them can recover the performance of the full model that uses all layers. However, we observe that the middle layers (13-16, 17-20) are slightly more useful than other layers.

\paragraph{Qualitative Study}

We show qualitative examples from XSum in Figure~\ref{fig:qual} to illustrate how \ours guided decoding improves performance. Greedy decoding from LLaMA-2-7B-Chat results in a hallucination, i.e. \textit{\$100m (£64m)}, that does not exist in the input document. However, the \ours is able to assign low scores for the chunk candidates that have contextual hallucinations (as marked in red). Therefore, \oursdecoding is able to help the model generate a summary that is factual to the given context.
\section{Related Work}
\paragraph{Hallucinations in LLMs}

\citet{simhi2024constructing} defined \emph{close-book hallucination} vs \emph{open-book hallucination} for settings of relying on parametric knowledge vs knowledge in context. We term \emph{open-book hallucination} as \emph{contextual hallucination} for better clarity. Previous studies in hallucinations primarily focus on close-book hallucinations~\cite{chen2023complex, min2023factscore, chern2023factool} and 
their detection~\citep{azaria2023internal, simhi2024constructing} and mitigation~\cite{li2024inference, chuang2023dola, chen2024context, zhang2024truthx}. Most of the studies focus on leveraging LLM's internal representations, such as hidden states~\citep{burns2022discovering, azaria2023internal}, MLP outputs~\citep{zhang2024truthx, simhi2024constructing}, attention block outputs~\citep{zhang2024truthx, simhi2024constructing} and attention head outputs~\citep{li2024inference, chen2024truth, simhi2024constructing}. Our work, however, focuses on contextual hallucinations, where models produce content inconsistent with the provided context~\cite{maynez2020faithfulness, fabbri2021qafacteval, shi2023trusting}. Thus, different from prior studies, we focus on the attention maps instead of internal representations, as we believe that the attention maps patterns record how the LLM process the given contextual information.
Most of the prior studies treat detection and mitigation as two separate tasks, expect for \citet{simhi2024constructing, chen2024context}. Our work focuses not only on detection, but also tries to incorporate the detector into the decoding process to further mitigate the contextual hallucinations.
Recently, \citet{simhi2024constructing} also explored detecting and mitigating both close-book and open-book hallucinations. However, their open-book hallucination setting is limited to DisentQA~\citep{neeman2023disentqa}, which creates knowledge conflicts between parametric knowledge and given context. In contrast, we focus on LLaMA-2's naturally generated responses to capture general cases where LLMs fail to follow the context, not just due to knowledge conflicts.

\paragraph{Classifier Guided Generation}

Classifier guided generation aims to control attributes like topic or sentiment in text generation. PPLM~\citep{dathathri2019plug} uses gradient ascent to adjust LM probabilities via attribute classifiers. FUDGE~\citep{yang2021fudge} uses an attribute predictor on partial sequences to modify LM probabilities. Our method uniquely guides generation using classifiers on attention maps, setting it apart from prior approaches.

\paragraph{Self-attention and Model Behavior}
The attention mechanism, initially introduced in RNN-based encoder-decoder for neural machine translation~\cite{bahdanau2015neural, luong2015effective}, was later adopted in the Transformer model's self-attention module~\cite{vaswani2017attention}, enabling greater parallelization. Self-attention's interpretability has led researchers to use it for understanding model behaviors~\cite{clark2019does,hao2021self,vashishth2019attention}. Our work demonstrates that attention maps in LLMs are effective for detecting contextual hallucinations, providing a lightweight and interpretable solution compared to complex hidden representation methods~\cite{zhang2024truthx, chen2024truth}.
\section{Conclusion}

We introduce the \ours, a lightweight classifier designed to detect contextual hallucinations by utilizing the \textit{lookback ratio}, which is computed solely from attention weights. This classifier not only effectively identifies contextual hallucinations but also mitigates them through \oursdecoding from the LLM. Remarkably, the method is transferable across various tasks, and even across models after mapping their attention heads. This research opens up new possibilities for leveraging attention map information to combat hallucinations in large language models.
\section*{Limitations}

Despite the effectiveness of the \ours and its decoding, there are several limitations to consider. 
\begin{itemize}
    \item First, the performance upper bound of \oursdecoding is limited by the sampling capabilities of the LLM itself. If the LLM fails to sample the correct chunk among the eight candidates, the \ours cannot correct the error. 
    \item Second, although the \ours is a lightweight classifier with negligible inference time, the requirement to sample multiple candidates from the LLM increases the total inference time. 
    We argue that \oursdecoding is a preliminary approach that demonstrates the feasibility of integrating the \ours into the decoding process, as well as a robustness test for the \ours to handle various text generation scenarios.
    However, other options, such as intervening in the attention map mechanism based on \ours signals, could potentially achieve faster inference, and we leave this for future work. 
    \item Lastly, the \ours relies on annotated examples of around 1k-2k to train the classifier. While other end-to-end methods~\citep{chuang2023dola} can mitigate close-book hallucinations without training data, they lack interpretability due to the absence of a detection step. Nevertheless, we believe that requiring 1,000 annotated examples is a feasible setting.
\end{itemize}

\section*{Acknowledgement}

We sincerely thank Philip Schroeder, Huirong Wen, Andrew Rouditchenko, Nishad Gothoskar, Ani Nrusimha, Howard Chen, Weijia Shi, and Nour Jedidi for their discussion and help in this project. This research was sponsored by the United States Air Force Research Laboratory and the United States Air Force Artificial Intelligence Accelerator and was accomplished under Cooperative Agreement Number FA8750-19-2-1000. The views and conclusions contained in this document are those of the authors and should not be interpreted as representing the official policies, either expressed or implied, of the United States Air Force or the U.S. Government. The U.S. Government is authorized to reproduce and distribute reprints for Government purposes notwithstanding any copyright notation herein. Linlu and Yoon were supported in part by MIT-IBM Watson AI Lab.

\section*{Ethics Statement}

In this research, we used publicly available datasets and we did not collect any personal information. 
All datasets and models are used in accordance with their intended use and licenses.
Our method is designed to improve the factuality of large language models (LLMs), which can have a positive impact on various applications, such as question-answering systems, summarization systems, and other applications that rely on LLMs. 
When deployed, however, our approach still carries the issues stemming from LLMs, which means that there is a risk that the LLM can produce biased, harmful, or offensive output.
Therefore, caution should be exercised before implementing similar approaches in real-world applications.

\newpage
\bibliography{main}

\newpage
\appendix
\section{Data Creation for \ours}
\label{appx:data_create}

Our experimental setup aims to evaluate the ability of \ours to detect hallucinations in large language models with attention maps. We consider the summarization task and question-answering (QA) task in data creation.

For the summarization task, we sampled 1,000 examples from the CNN/DM dataset~\citep{see2017get}.
For QA, we use 2,655 examples from the Natural Questions~\citep{kwiatkowski2019natural} from the setup of \citet{liu2024lost} to mix the gold document with irrelevant documents. To keep our focus more on LLM hallucinations rather than being distracted by assessing LLMs' long-context utilization ability, we limited context to three documents per question where the gold document containing the answer was placed in the middle, surrounded by two irrelevant documents. 

We prompt LLaMA-2-7B-Chat~\citep{touvron2023llama} to generate correct responses by greedy decoding for both tasks to ensure that both hallucinated and non-hallucinated examples derive from the same source distribution. The max length of generation is set to 256 tokens, or until the EOS token is generated.

After the annotation was collected, we extract hallucinated and non-hallucinated spans, as well as the corresponding attention map lookback ratio, from the LLaMA-2-7B-Chat model, to train the \ours classifiers.

In the predefined span setting, three types of spans are considered as non-hallucinated spans: 1) the text segment before the first hallucinated span in the response 2) the text segment after the last hallucinated span in the response 3) the response annotated as non-hallucinated. All the annotated hallucinated spans are used as negative data to train the \ours.

In the sliding window setting, we consider all the possible fixed sized chunk with size = 8. If a chunk is overlapping with any of the annotated hallucinated spans, then it is considered as hallucinated, otherwise it is non-hallucinated.

\paragraph{Why not use existing data?} Initially, we considered using the HaluEval dataset~\citep{li2023halueval}, which was created by prompting GPT-3.5~\citep{chatgpt2023} to generate ``hallucinated examples'' against human-annotated non-hallucinated responses, on summarization, QA, and dialogue tasks. However, we have concerns that their method introduces a bias by creating fundamentally different data distributions between hallucinated and non-hallucinated examples. This discrepancy could potentially lead the classifier to learn to distinguish the sources of responses rather than accurately detecting hallucinations. 

Additionally, we argue that the LLM's attention weight will be more meaningful if the text is generated by the same LLM itself, not from external sources and teacher forcing to obtain the attention weights.
To ensure an unbiased and controlled evaluation environment, we generated our own dataset on summarization and QA tasks. 
\section{Evaluation Details}
\subsection{Evaluation Prompt for GPT-4o}
\label{appx:gpt4o}
We show the templates used to prompt GPT-4o (\texttt{gpt-4o-2024-05-13}) in annotating the truthfulness of a response and the span-level hallucination segment prediction in Table~\ref{tab:gpt4-prompt-sum} and Table~\ref{tab:gpt4-prompt-nq}, respectively for CNN/DM and Natural Questions.

This prompt is used for 1) collecting the data to train the \ours in Table~\ref{tab:dataset_stats}, and 2) evaluating the XSum summarization task in Sections~\ref{sec:mitigate}, \ref{sec:cross}, and \ref{sec:discuss}.
We also provide the approximate cost of GPT-4o calls (in USD):
\begin{itemize}
    \setlength\itemsep{0em}
    \item 1000 examples from XSum is around \$8.
    \item 1000 examples from CNN/DM is around \$12.
    \item 2655 examples from NQ is around \$16.
\end{itemize}

\begin{table*}[h!]
\centering
\small
\begin{tabular}{p{\dimexpr \linewidth-2\tabcolsep}}
\toprule
\begin{minipage}[t]{\linewidth}
\texttt{You will be provided with a document and a proposed summary. Your task is to determine if the proposed summary can be directly inferred from the document. If the summary contains any information not found in the document, it is considered false. Even if the summary is different from a ground truth summary, it might still be true, as long as it doesn't contain false information.\\
For each proposed summary, explain why it is true or false based on the information from the document. Focus only on the original document's content, disregarding any external context.\\
After your explanation, give your final conclusion as \red{Conclusion: True} if the proposed summary is completely accurate based on the document, or \red{Conclusion: False} if it contains any incorrect or unsupported information. If your conclusion is 'False', identify the exact phrases or name entities from the summary that is incorrect by stating \red{Problematic Spans: [the inaccurate text spans from the summary, in Python list of strings format]}.\\
\\
\#Document\#: \{document\}\\
\\
\#Ground Truth Summary\#: \{ground\_truth\_summary\}\\
\\
\#Proposed Summary\#: \{response\}\\
\\
Write your explanation first, and then give your final conclusion as \red{Conclusion: True} if the proposed summary is completely accurate based on the document, or \red{Conclusion: False} if it contains any incorrect or unsupported information. Add \red{Problematic Spans: [the exact inaccurate text spans from the summary, in a list of strings]} if your conclusion is 'False'.}
\end{minipage} \\
\bottomrule
\end{tabular}
\caption{Prompt template for GPT-4o in annotating the truthfulness and predicting span-level hallucinations on summarization tasks. Used for CNN/DM and XSum.}
\label{tab:gpt4-prompt-sum}
\end{table*}

\begin{table*}[h!]
\centering
\small
\begin{tabular}{p{\dimexpr \linewidth-2\tabcolsep}}
\toprule
\begin{minipage}[t]{\linewidth}
\texttt{You will be provided with a document and a proposed answer to a question. Your task is to determine if the proposed answer can be directly inferred from the document. If the answer contains any information not found in the document, it is considered false. Even if the answer is different from a ground truth answer, it might still be true, as long as it doesn't contain false information.\\
For each proposed answer, explain why it is true or false based on the information from the document. Focus only on the original document's content, disregarding any external context.\\
After your explanation, give your final conclusion as \red{Conclusion: True} if the proposed answer is completely accurate based on the document, or \red{Conclusion: False} if it contains any incorrect or unsupported information. If your conclusion is 'False', identify the exact phrases or name entities from the answer that is incorrect by stating \red{Problematic Spans: [the inaccurate text spans from the answer, in Python list of strings format]}.\\
\\
\#Document\#: \{document\}\\
\\
\#Ground Truth Answers (a list of valid answers)\#: \{ground\_truth\_answers\}\\
\\
\#Proposed Answer\#: \{response\}\\
\\
Write your explanation first, and then give your final conclusion as \red{Conclusion: True} if the proposed answer is completely accurate based on the document, or \red{Conclusion: False} if it contains any incorrect or unsupported information. Add \red{Problematic Spans: [the exact inaccurate text spans from the answer, in a list of strings]} if your conclusion is 'False'.}
\end{minipage} \\
\bottomrule
\end{tabular}
\caption{Prompt template for GPT-4o in annotating the truthfulness and predicting span-level hallucinations on question-answering tasks. Used for Natural Questions.}
\label{tab:gpt4-prompt-nq}
\end{table*}

\subsection{Human Evaluation on GPT-4o Evaluation}
\label{appx:human}

\paragraph{Summarization} 

To assess the quality of GPT-4o’s evaluations, we initially conducted a pilot study using 70 XSum dataset examples, with native English-speaking authors and colleagues as evaluators. Evaluators received the document, ground truth summary, LLaMA-2-7B-Chat’s summary, and GPT-4o’s judgment to provide a binary judgment on GPT-4o’s accuracy. Our interface is depicted in Appendix~\ref{appx:gpt4o} (see Figure~\ref{fig:humananno}). This initial evaluation affirmed the correctness of GPT-4o’s judgments in 68 out of 70 cases. To further verify these results, we expanded our evaluation through Amazon MTurk, adding two additional annotations per example. Across all 210 evaluations (70 initial + 140 MTurk), only 9 annotations were marked incorrect, and in only 2 cases did a majority of annotators deem the judgment incorrect (marked incorrect by at least two annotators). With a final accuracy of 97.1\%, and high intra-annotator agreement, the comprehensive evaluation supports GPT-4o’s use as an automatic evaluator for the entire dataset.

\paragraph{Question Answering}

We expand the human evaluation to Natural Questions dataset using Amazon MTurk. The evaluation interface is copied from the summarization setup, but changing “summary” to “answer”, as well as adding the “question” field.

We take 50 examples and assign each example to three different annotators. There are 7 annotations marked incorrect out of the 150 annotations. In total, 3 of the examples are marked incorrect by at least two annotators. If applying a majority vote, 47 out of 50 examples are correct, resulting in a 94.0\% accuracy. This suggests that it is generally sufficient to use GPT-4o to verify the generated responses on the question-answering task.

\begin{figure}[t!]
\begin{center}
\includegraphics[width=\columnwidth]{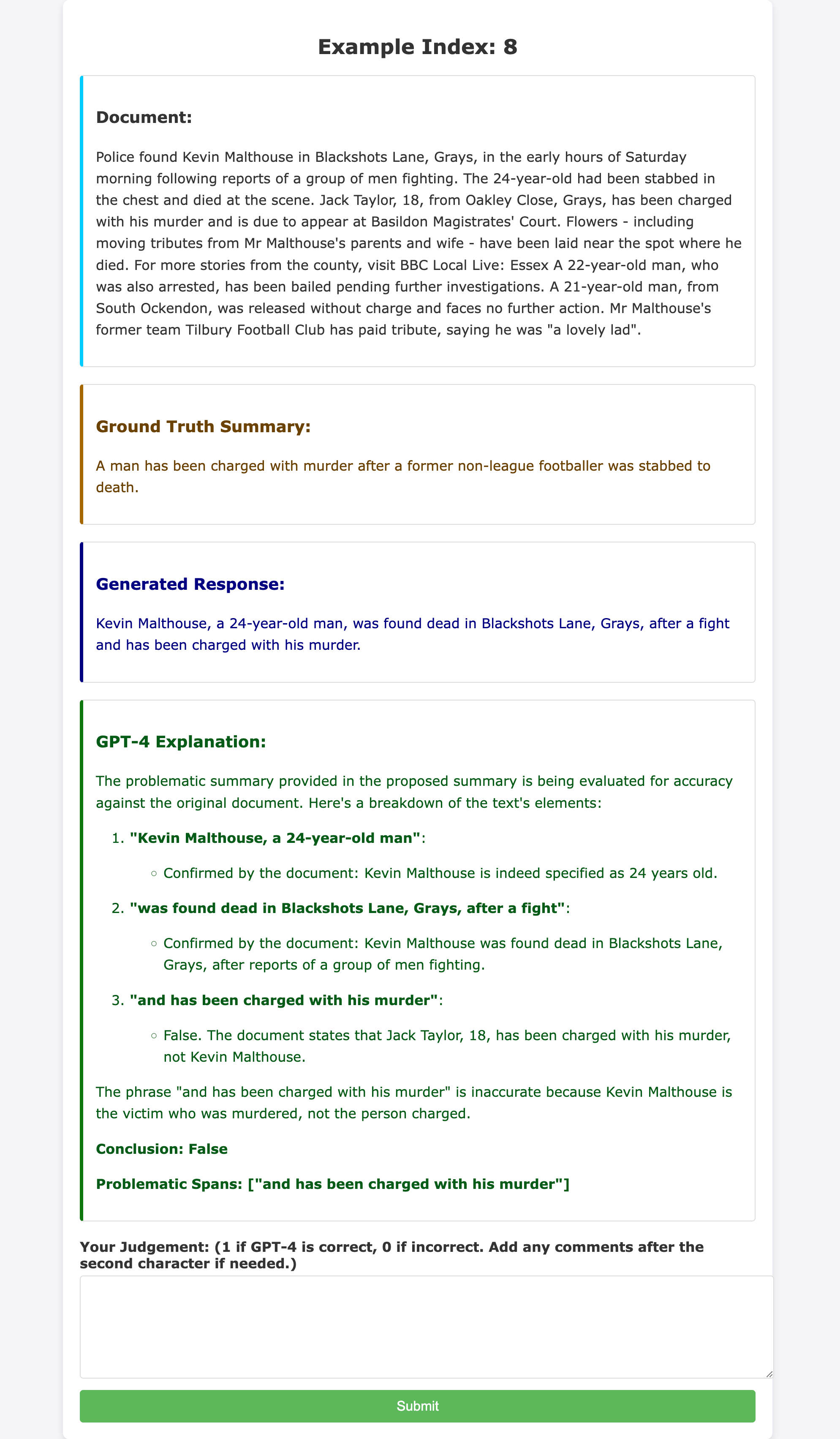}
\end{center}
\caption{Screenshot of human annotation interface.}
\label{fig:humananno}
\end{figure}

\subsection{Evaluation Prompt for MT-Bench}

We show the evaluation prompt for MT-Bench (hallucination) in Table~\ref{tab:mt-bench-hallu-prompt}. We follow standard practice for MT-Bench (original) evaluation\footnote{\url{https://github.com/lm-sys/FastChat/tree/main/fastchat/llm_judge}.} and show evaluation prompts in Table~\ref{tab:mt-bench-general-prompt}. We evaluate MT-bench (original) with their default GPT-4 model \texttt{gpt-4-0613} and our proposed MT-Bench (hallucination) with the latest GPT-4o model (\texttt{gpt-4o-2024-05-13}).

\label{appx:mtbench}
\begin{table*}[h!]
\centering
\small
\begin{tabular}{p{\dimexpr \linewidth-2\tabcolsep}}
\toprule
\begin{minipage}[t]{\linewidth}
\texttt{Please act as an impartial judge and evaluate the \textcolor{red}{faithfulness and consistency} of the response provided by an AI assistant to the user question displayed below. Your evaluation should consider \textcolor{red}{whether the assistant's answer to the second user question is faithful and consistent to the chat history. If the answer contains any misinformation not found or not supported by the chat history, it is considered a hallucination.} You evaluation should focus on the assistant's answer to the second user question. Begin your evaluation by providing a short explanation. Be as objective as possible. After providing your explanation, you must rate the response on a scale of 1 to 10 by strictly following this format: ``[[rating]]", for example: ``Rating: [[5]]". \\
\\
<|The Start of Assistant A's Conversation with User|> \\
\\
\#\#\# User:\\
\{question\_1\}\\
\\
\#\#\# Assistant A:\\
\{answer\_1\}\\
\\
\#\#\# User:\\
\{question\_2\}\\
\\
\#\#\# Assistant A:\\
\{answer\_2\} \\
\\
<|The End of Assistant A's Conversation with User|>}
\end{minipage} \\
\bottomrule
\end{tabular}
\caption{GPT-4o evaluation prompt for MT-bench (hallucination).}
\label{tab:mt-bench-hallu-prompt}
\vspace{-5pt}
\end{table*}

\begin{table*}[h!]
\centering
\small
\begin{tabular}{p{\dimexpr \linewidth-2\tabcolsep}}
\toprule
\begin{minipage}[t]{\linewidth}
\texttt{Please act as an impartial judge and evaluate the \textcolor{red}{quality} of the response provided by an AI assistant to the user question displayed below. Your evaluation should consider \textcolor{red}{factors such as the helpfulness, relevance, accuracy, depth, creativity, and level of detail of the response}. You evaluation should focus on the assistant's answer to the second user question. Begin your evaluation by providing a short explanation. Be as objective as possible. After providing your explanation, you must rate the response on a scale of 1 to 10 by strictly following this format: "[[rating]]", for example: "Rating: [[5]]".\\
\\
<|The Start of Assistant A's Conversation with User|>\\
\\
\#\#\# User:\\
\{question\_1\}\\
\\
\#\#\# Assistant A:\\
\{answer\_1\}\\
\\
\#\#\# User:\\
\{question\_2\}\\
\\
\#\#\# Assistant A:\\
\{answer\_2\}\\
\\
<|The End of Assistant A's Conversation with User|>}
\end{minipage} \\
\midrule
\begin{minipage}[t]{\linewidth}
\texttt{Please act as an impartial judge and evaluate the \textcolor{red}{quality} of the response provided by an AI assistant to the user question. Your evaluation should consider \textcolor{red}{correctness and helpfulness. You will be given a reference answer and the assistant's answer}. You evaluation should focus on the assistant's answer to the second question. Begin your evaluation by comparing the assistant's answer with the reference answer. Identify and correct any mistakes. Be as objective as possible. After providing your explanation, you must rate the response on a scale of 1 to 10 by strictly following this format: "[[rating]]", for example: "Rating: [[5]]".\\
\\
<|The Start of Reference Answer|>\\
\\
\#\#\# User:\\
\{question\_1\}\\
\\
\#\#\# Reference answer:\\
\{ref\_answer\_1\}\\
\\
\#\#\# User:\\
\{question\_2\}\\
\\
\#\#\# Reference answer:\\
\{ref\_answer\_2\}\\
\\
<|The End of Reference Answer|>\\
\\
\\
<|The Start of Assistant A's Conversation with User|>\\
\\
\#\#\# User:\\
\{question\_1\}\\
\\
\#\#\# Assistant A:\\
\{answer\_1\}\\
\\
\#\#\# User:\\
\{question\_2\}\\
\\
\#\#\# Assistant A:\\
\{answer\_2\}\\
\\
<|The End of Assistant A's Conversation with User|>\\
}
\end{minipage} \\
\bottomrule
\end{tabular}
\caption{GPT-4 evaluation prompt for general questions (top) and math questions (bottom) on MT-bench (original).}
\label{tab:mt-bench-general-prompt}
\end{table*}

\section{Experiment Details}
\subsection{Model Details}
\label{appx:nli}

\paragraph{State-of-the-art NLI Model} We give further detail on the pretrained SoTA NLI model~\footnote{\url{https://huggingface.co/vectara/hallucination_evaluation_model}} used as our topline hallucination detector. Specifically, the model is based on DeBERTa-V3-base~\cite{he2021debertav3} and further finetuned on a range of NLI and summarization datasets with examples annotated with factual consistency, including FEVER~\citep{Thorne18Fever}, Vitamin C~\citep{schuster-etal-2021-get} and PAWS~\citep{paws2019naacl}. Roughly 731k data examples can be collected from the training set of the above three datasets.
The model is reported to have superior performance when evaluated on TRUE~\citep{honovich2022true} SummaC Benchmark~\citep{laban2022summac} and AnyScale Ranking Test for Hallucinations~\footnote{\url{https://www.anyscale.com/blog/llama-2-is-about-as-factually-accurate-as-gpt-4-for-summaries-and-is-30x-cheaper}}.

\paragraph{Other Model Details and License}

\begin{itemize}
    \item $\texttt{Llama-2-7B-Chat}$: A 7B parameter model that is instruction fine-tuned. HuggingFace ID: $\texttt{meta-llama/Llama-2-7b-chat-hf}$.
    \item $\texttt{Llama-2-13B-Chat}$: A 13B parameter model that is instruction fine-tuned. HuggingFace ID: $\texttt{meta-llama/Llama-2-13b-chat-hf}$.
    \item  $\texttt{hallucination\_evaluation\_model}$: Based on $\texttt{microsoft/deberta-v3-base}$ which has 86M parameters. HuggingFace ID: $\texttt{vectara/hallucination\_evaluation\_model}$.
    \item  $\texttt{DeBERTa-V3-Base}$: a 86M parameters encoder based model. HuggingFace ID: $\texttt{microsoft/deberta-v3-base}$.
\end{itemize}

The above models have the following licenses. 
\begin{itemize}
    \item $\texttt{Llama-2-7B-Chat}$ is under the Llama 2 Community License Agreement. 
    \item $\texttt{Llama-2-13B-Chat}$ is under the Llama 2 Community License Agreement. 
    \item $\fontsize{10.4}{10}\texttt{vectara/hallucination\_evaluation\_model}$ \normalsize is under the Apache 2.0 License. 
    \item $\texttt{DeBERTa-V3-Base}$ is under MIT License.
\end{itemize}

\paragraph{Inference Details}
We run all the models on NVIDIA A6000 (48GB) and V100 (32GB) GPUs. We do not train the model, but only run the inference part. Each of the examples takes around 20-30 seconds for 7B model, 40-60 seconds for 13B model to generate responses using our \oursdecoding. Please check Appendix~\ref{appx:data} to estimate the total running time on each of the datasets, as it depends on number of examples.

All the inferences are run with either greedy decoding or sampling using temperature 0.9 and top-$p$ sampling with $p=0.95$. The implementation is based on Huggingface Transformers packages.\footnote{\url{https://github.com/huggingface/transformers}}
All the scores in the paper are from a single run due to the limited computation for the large models.
\paragraph{Classifier Training Details}
We use Scikit-Learn \texttt{sklearn.linear\_model.LogisticRegression}\footnote{
\url{https://scikit-learn.org/stable/modules/generated/sklearn.linear_model.LogisticRegression.html}} to train the classifiers of \ours on CPU machine. We use all the default hyperparameters, such as L2 penalty, etc, but we change the \texttt{max\_iter} to 1000 to ensure it is converged.

\paragraph{Heads Mapping Details}
We use Scikit-Learn \texttt{sklearn.linear\_model.LinearRegression}\footnote{
\url{https://scikit-learn.org/stable/modules/generated/sklearn.linear_model.LinearRegression.html}} in Section~\ref{sec:cross}, to fit a linear transformation from LLaMA-2-13B-Chat's attention heads to LLaMA-2-7B-Chat's attention heads. It is computed to solve the close-form Ordinary Least Squares optimization problem, without gradient descent. We use all the default hyperparameters and run it on our CPU machine.

\subsection{Dataset Details}
\label{appx:data}

The datasets we used in the paper have the following details:

\begin{itemize}
\item CNN/DM: sampled 1000 examples from the testing set. Apache-2.0 license. \url{https://huggingface.co/datasets/abisee/cnn_dailymail}
\item Natural Questions: Apache-2.0 license. Testing set: 2655 examples from \url{https://github.com/nelson-liu/lost-in-the-middle}. NQ-train: sampled 2499 examples from its training set, using the positive document provided by \url{https://github.com/facebookresearch/DPR}
\item XSum: 1000 examples sampled from the testing set. MIT license. \url{https://github.com/EdinburghNLP/XSum}
\item MT-bench: 80 examples. Apache-2.0 license. \url{https://github.com/lm-sys/FastChat/tree/main/fastchat/llm_judge}
\end{itemize}

\section{Additional Results}
\subsection{Visualization}
\label{appx:vis}
\begin{figure}[t!]
\begin{center}
\includegraphics[width=\columnwidth]{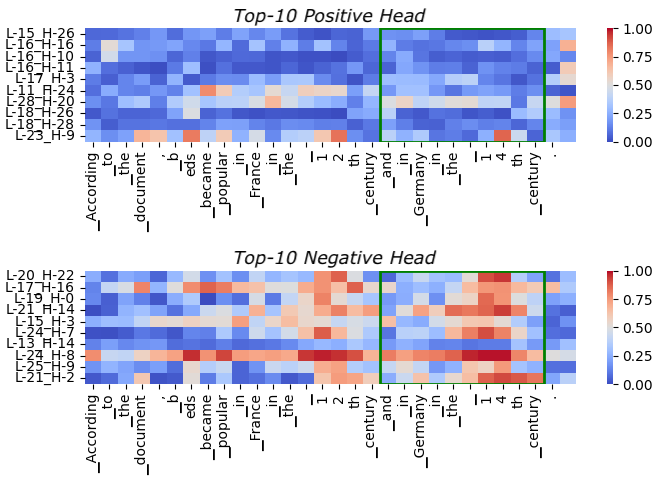}
\end{center}
\caption{Top-10 positive/negative heads ranked from top to the bottom by the magnitude of their coefficients in the \ours classifier.}
\label{fig:vis}
\end{figure}

We visualize the lookback ratio of the top-10 most positive/negative heads when LLaMA-2-7B-Chat decodes the answer for an NQ example. 
The top-10 most positive/negative heads are selected with the most positive/negative coefficients from the classifier.
The green rectangle frames the part that contains the hallucinations, i.e. \textit{and in Germany in the 14th century}.
We can see that during the generation of the hallucinated span, the positive heads, especially for the top-1 heads (topmost), show a lower lookback ratio (in blue), while the negative heads show a slightly higher lookback ratio (in red). However, the behavior of \ours still needs to be determined by the collective behavior of all heads and the weight and bias of the classifier.

\subsection{Using Multiple or All Layers for Hidden States}
\label{appx:all_layers}

\paragraph{Multiple Layer} We follow the prior study~\citep{azaria2023internal} to use the layers with the best predictive power in hallucination detection: 32nd/28th/24th/20th layers. We concatenate the 4 layer features into a huge feature. Please note that the hidden dimension of LLaMA-7B is 4096, so combining 4 layers would result in a 16384-dim feature vector. In contrast, our \ours feature for the 7B model is only 1024-dim. Thus, the big classifier using 16384 input features is supposed to be more effective given that it uses 10x more features.

However, the result shown in Table~\ref{tab:all_layers} indicates that concatenating 4 layers is still less effective compared to our Lookback Lens.

\paragraph{All Layers} We also try to use the hidden states from all layers, but concatenating them all will result in a huge feature vector with dimensions of more than 100k and make the classifier extremely slow in training. Thus, we perform max/average pooling for the features across different layers, resulting in 4096-dim feature vectors as the classifier inputs. The results shown in the table below are still worse than our \ours results.

The two experiments above indicate that using multiple or all layers may not be the key to making the classifier accurate. Instead, by designing good features like lookback ratio, the compact 1024-dim feature can be even more effective compared to the 10x bigger high-dimensional hidden state features.

\begin{table}[t!]
    \centering
    \small
    \begin{tabular}{lccc}
        \toprule
        \multirow{3}{*}{\bf Method} & \multicolumn{3}{r}{\bf AUROC (sliding window = 8)} \\
        \cmidrule(l{5em}r){2-4}
        & \multicolumn{2}{r}{\bf NQ $\rightarrow$ Sum.} & \bf Sum. $\rightarrow$ NQ \\
        \midrule
        \multicolumn{4}{l}{\it Residual outputs (hidden states)} \\
        \multicolumn{2}{l}{Layer 32} & 56.1 & 59.4 \\
        \multicolumn{2}{l}{Layer 28} & 57.7 & 58.8 \\
        \multicolumn{2}{l}{Layer 24} & 58.3 & 58.3 \\
        \multicolumn{2}{l}{Layer 20} & 57.6 & 59.5 \\
        \multicolumn{2}{l}{Concatenate above 4 layers} & 58.8 & 59.2 \\
        \multicolumn{2}{l}{Max pooling all 32 layers} & 56.7 & 59.2 \\
        \multicolumn{2}{l}{Average pooling all 32 layers} & 57.3 & 59.2 \\
        \midrule
        \multicolumn{2}{l}{Ours: Lookback Lens} & 66.1 & 66.0 \\
        \bottomrule
    \end{tabular}
    \caption{AUROC results for different methods of utilizing hidden states.}
    \label{tab:all_layers}
\end{table}

\subsection{Comparing Attention Outputs with Hidden States}
\label{appx:attn-out}

Some papers mention that attention block outputs could be more useful for detecting hallucinations~\citep{campbell2023localizing, li2024inference}, while our main experiments only consider the hidden states as input features for detecting contextual hallucinations. Here we include additional experiment results that use attention block outputs instead. In Table~\ref{tab:attn-out}, we show that there is no significant difference when switching to attention block outputs, and our \ours still outperforms these baselines.

\begin{table}[t!]
    \centering
    \small
    \begin{tabular}{lcc}
        \toprule
        \multirow{3}{*}{\bf Method} & \multicolumn{2}{c}{\bf AUROC (sliding window = 8)} \\
        \cmidrule(lr){2-3}
        & \bf NQ $\rightarrow$ Sum. & \bf Sum. $\rightarrow$ NQ \\
        \midrule
        \multicolumn{3}{l}{\it Attention block outputs} \\
        Layer 32 & 57.6 & 60.7 \\
        Layer 28 & 58.5 & 57.2 \\
        Layer 24 & 56.3 & 57.2 \\
        \midrule
        \multicolumn{3}{l}{\it Residual outputs (hidden states)} \\
        Layer 32 & 56.1 & 59.4 \\
        Layer 28 & 57.7 & 58.8 \\
        Layer 24 & 58.3 & 58.3 \\
        \midrule
        Ours: Lookback Lens & 66.1 & 66.0 \\
        \bottomrule
    \end{tabular}
    \caption{AUROC results for different layers and outputs.}
    \label{tab:attn-out}
\end{table}

\end{document}